\documentclass[10pt,twocolumn,letterpaper]{article}

\usepackage{cvpr}

\usepackage{amsmath,amsfonts,bm}

\def\figref#1{figure~\ref{#1}}

\def\secref#1{section~\ref{#1}}

\def\eqref#1{equation~\ref{#1}}

\def\1{\bm{1}}

\def\vc{{\bm{c}}}
\def\vd{{\bm{d}}}

\def\vp{{\bm{p}}}

\def\vr{{\bm{r}}}

\def\vv{{\bm{v}}}

\def\vx{{\bm{x}}}

\def\mD{{\bm{D}}}

\def\mI{{\bm{I}}}

\def\mP{{\bm{P}}}

\def\mV{{\bm{V}}}

\DeclareMathAlphabet{\mathsfit}{\encodingdefault}{\sfdefault}{m}{sl}
\SetMathAlphabet{\mathsfit}{bold}{\encodingdefault}{\sfdefault}{bx}{n}

\usepackage[utf8]{inputenc} 
\usepackage[T1]{fontenc}    
\usepackage[pagebackref=true,breaklinks=true,letterpaper=true,colorlinks,bookmarks=false]{hyperref}      
\usepackage{url}            
\usepackage{booktabs}       
\usepackage{amsfonts}       
\usepackage{nicefrac}       
\usepackage{microtype}
\usepackage{hhline}
\usepackage{xcolor}
\usepackage{color}
\usepackage{times}
\usepackage{epsfig}
\usepackage{graphicx}
\usepackage{amsmath}
\usepackage{amssymb}
\usepackage{amsthm}
\usepackage{array}
\usepackage{multirow}
\usepackage{booktabs}
\usepackage{subfigure}
\usepackage{float}
\usepackage[numbers,square]{natbib}

\renewcommand{\figref}[1]{Fig.~\ref{fig:#1}}

\newcommand{\equref}[1]{Eq.~(\ref{equ:#1})}
\renewcommand{\secref}[1]{Sec.~\ref{sec:#1}}

\newcommand{\myinput}[1]{}
\renewcommand{\cite}[1]{\citet{#1}}
\renewcommand{\paragraph}[1]{\noindent{\bf{#1}}}

\usepackage{caption}
\definecolor{myblue}{HTML}{004D7F}
\definecolor{myred}{HTML}{B51700}
\hypersetup{colorlinks,linkcolor={myred},citecolor={myblue}}

\cvprfinalcopy 
\def\cvprPaperID{1515} 

\ifcvprfinal\fi
\begin{document}

\title{LSM: Learning Subspace Minimization for Low-level Vision}
\twocolumn[{%
\maketitle
\vspace{-1.5cm}
\begin{center}
{
\vspace{-0.5cm}
\normalsize
Chengzhou Tang$^{1}$~~~~~~Lu Yuan$^{2}$~~~~~~Ping Tan$^1$\\\vspace{0.1cm}
$^1$Simon Fraser University~~~~~~$^2$Microsoft
}
\centering
\vspace{0.3cm}
\includegraphics[width=0.9\linewidth]{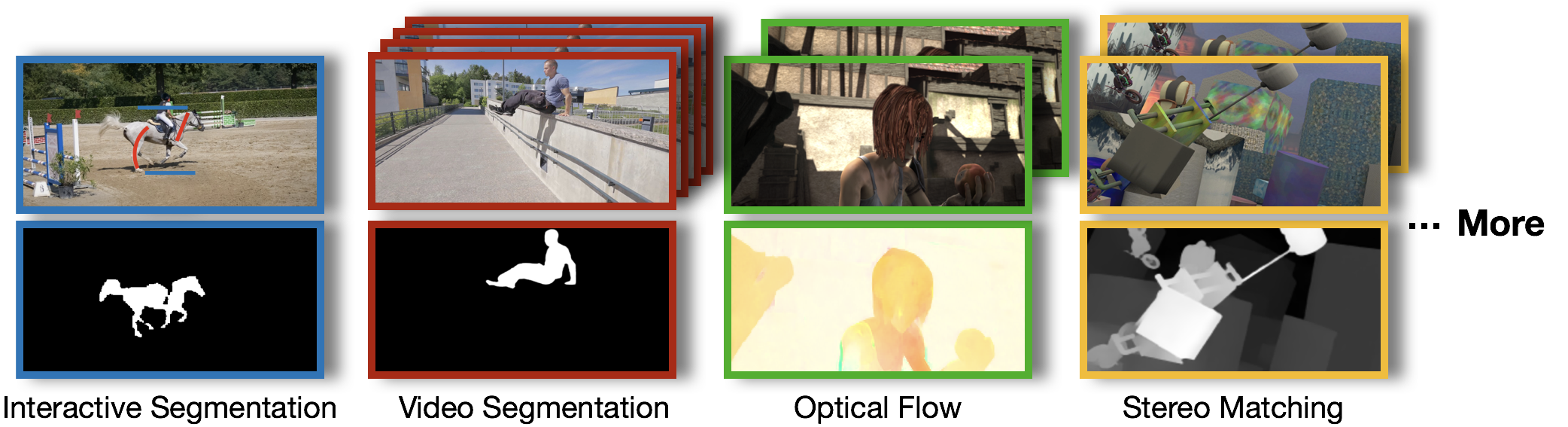}
\vspace{-10pt}
    \captionof{figure}{Learning subspace minimization solves various low-level vision tasks using \emph{unified} network structures and parameters.}\label{fig:teaser}
\end{center}%
}]

\begin{abstract}
We study the energy minimization problem in low-level vision tasks from a novel perspective.
We replace the heuristic regularization term with a learnable subspace constraint, and preserve the data term to exploit domain knowledge derived from the first principle of a task.
This learning subspace minimization (LSM) framework unifies the network structures and the parameters for many low-level vision tasks, which allows us to train a single network for multiple tasks simultaneously with completely shared parameters, and even generalizes the trained network to an unseen task as long as its data term can be formulated. 
We demonstrate our LSM framework on four low-level tasks including interactive image segmentation, video segmentation, stereo matching, and optical flow, and validate the network on various datasets. The experiments show that the proposed LSM generates state-of-the-art results with smaller model size, faster training convergence, and real-time inference. 
\end{abstract}
\section{Introduction}
\label{sec:intro}
Many low-level vision tasks (\eg image segmentation~\citep{IMGSEG,TVSEG,Gulshan10}, video segmentation~\citep{BVS,OFL,DAC}, stereo matching~\citep{VARSTEREO1,VARSTEREO2,VARSTEREO3} and optical flow~\citep{HS,FLOWARP,Sun2014}) are conventionally formulated as an energy minimization problem:
\vspace{-5pt}
\begin{equation}
\min_{\vx}D(\vx)+R(\vx),
\label{equ:1}
\vspace{-5pt}
\end{equation}
where $\vx$ is the desired solution (\eg a disparity field for stereo matching), and the two terms $D(\vx)$ and $R(\vx)$ are the data term and regularization term respectively. The data term $D(\vx)$ is usually well designed following the first principle of a task, such as the color consistency assumption in stereo and optical flow.
However, the regularization term $R(\vx)$ is often heuristic. Typically, it regularizes $\vx$ at the~\emph{pixel-level} and encourages similar pixels to have similar solution values. 
The regularization term is necessary because low-level vision tasks are usually ill-posed~\citep{ILLPOSE}, and a standalone data term is often insufficient,~\eg the aperture problem in optical flow. 

However, a vanilla $L^2$ smoothness regularization~\citep{TAREG} may cause over-smoothed results at object boundaries. Ideally, the regularization term should smooth out noises in $\vx$ and preserve sharp edges. Thus, many edge-preserving regularization terms have been developed, such as the Total Variation (TV) regularization~\citep{ROF,TVINTRO}, the anisotropic diffusion~\citep{PM}, the bilteral filter~\citep{BL} which focuses on designing better similarity measurements between pixels, and the distance in a learned feature embedding space~\citep{ROTH2009,LEARNFLO,SPN} has also been adopted for the same purpose. But it is still an unsolved problem to design an ideal similarity measurement for efficient and accurate energy minimization.

We study this energy minimization problem from a different perspective. Instead of focusing on the \emph{pixel-level} similarity, we exploit \emph{image-level} context information. Specifically, we preserve the data term $D(\vx)$ but replace the heuristic regularization term $R(\vx)$ with a subspace constraint:
\vspace{-5pt}
\begin{equation}
\min_{\vx}D(\vx), \text{~s.t.~} \vx\in\mathcal V=\text{span}\{\vv_{1},\cdots\vv_{K}\},
\label{equ:2}
\vspace{-5pt}
\end{equation}
where $\mathcal V$ is a $K$-dimensional subspace, and~$\{\vv_{1},\cdots\vv_{K}\}$ is the corresponding basis vectors.
Our motivation is different from the regularization term $R(\vx)$: we use the~\emph{image-level} context information to regularize the problem by assuming the desired solution $\vx$ is composited of several layers~\citep{LAYER1,LAYER2,LAYER3},~\eg motion layers for optical flow, and each basis vector $\vv_k$ will correspond to one of these layers. Therefore, we can represent the solution $\vx$ as a linear combination of these basis vectors and solve the combination coefficients, leading to a compact minimization that~\emph{not only} is efficient~\emph{but also} enables end-to-end training and outperforms the conventional regularization term $R(\vx)$.

To this end, we propose the \emph{learning subspace minimization} (LSM) framework that progressively evolves $\mathcal V$ and solves~\equref{2} on a feature pyramid coarse-to-fine.
At each pyramid level, we employ a convolutional neural network (CNN) to update $\mathcal V$ from both the image features and the derivatives of the data term $D(\vx)$ respect to the intermediate solution $\vx$. 
Since the generation of $\mathcal V$ receives the task-specific data term as the~\emph{input}, it decouples the task-specific characteristics from the subspace generation and unifies the network structures and the parameters for various tasks. 

As a consequence, our LSM framework enables joint multi-task learning with~\emph{completely} shared network structures as well as parameters, and even makes zero-shot task generalization possible, where a trained network is plug-and-play for an unseen task without any parameter modification, as long as the corresponding data term $D(\vx)$ is formulated.

In the experiments, we implement four low-level vision tasks in an unified paradigm, including interactive image segmentation, video segmentation, stereo matching and optical flow. Our LSM framework has achieved better or comparable results with state-of-the-art methods. Our network structures and parameters can be unified into a compact model, which yields higher efficiency in training and inference.
We also demonstrate zero-shot task generalization by leaving one task out for testing and train on the other tasks.
All these benefits come from our methodology that integrates domain knowledge (\ie minimizing a data term derived from the first principle) with convolutional neural networks (\ie learning to generate subspace constraint).
\section{Related Works}
\paragraph{Regularization in Variational Method} 
Many computer vision problems can be formulated to~\equref{1}.
We only review the continuous settings (\ie variational method) because it is more relevant to our work and refer readers to~\citep{kohli2012higher-order} for the review about the discrete settings.
One of the main focuses of these works is on designing appropriate objective function, especially the regularization term.
\cite{ROF} first proposed the TV regularization for image denoising, which has also been proven to be successful for image super-resolution~\citep{TVSUP}, interactive image segmentation~\citep{TVSEG}, stereo matching~\citep{VARSTEREO1}, optical flow~\citep{TVFLOW1,TVFLOW2}, multi-view stereo~\citep{TVMV},~\emph{etc}.~\citet{PM} pioneered to use partial differential equations (PDE) for anisotropic diffusion, which is equivalent to minimizing an energy function with edge-aware regularization~\citep{EDGEAW1,BL}. Non-local regularizations~\citep{nonlocals2} have also been proposed for image super-resolution~\citep{nonlocalsup}, image inpainting~\citep{nonlocal2}, optical flow~\citep{NONLOCALS},~\emph{etc}, which performs better by connecting longer range pixels but is usually computational expensive. 

Our LSM framework also minimizes an objective function. But we only preserve the data term since it is usually derived from the first principle of a task, and replace the heuristic regularization term to a learned subspace constraint that captures the structure of the desired solution at the whole image context level and enables end-to-end training to boost the performance from data.

\paragraph{Convolutional Neural Networks} 
Inspired by the success of CNNs in high-level tasks~\citep{ALEXNET,VGG,RESNET}, numerous CNN based methods have been proposed for low-level vision tasks.~\cite{SRCNN} pioneered to use a CNN to upsample image patches for super-resolution.
\cite{zbontar2016stereo} and~\cite{7780983} used CNN features to measure image patches' similarity for stereo matching,~\cite{XRK2017} and~\cite{FF} also used CNN based similarity for optical flow. All these methods used CNNs in the patch level, which is computationally expensive and requires post-processing to composite the final result. So more recent works used whole images as inputs.~\cite{FLOWNET} used an encoder-decoder structure for optical flow, which is then extended to stereo matching~\citep{SYNDATA} and further evolved in~\cite{FLOWNET2,FLOWNET3} and other works~\citep{PWCNET,PSMNET,Yin_2019_CVPR}. Some recent works~\citep{DIOS,LATENT,BRS} enabled interactive image segmentation by feeding an image and an user annotation map to CNNs. Meanwhile, CNN based methods~\citep{MASKRNN,VIDEOMATCH,FEELVOS,MASKTRACK} have also achieved leading performance for video segmentation. 

Our LSM framework also employs CNNs but for different purposes. Instead of predicting the solution directly, we use CNNs to constraint the solution onto a subspace to facilitate the minimization of the data term. The data term is derived from the first principle of each task and decouples the task-specific formulation from the network parameters. Therefore, our framework unifies the network structures as well as parameters for different tasks and even enables zero-shot task generalization, which are difficult for fully CNN based methods. Although some recent works~\citep{CODESLAM,BA} also learn to generate subspace via CNNs, they are designed specifically for 3D reconstruction, which is ad-hoc and unable to generalize to broader low-level vision tasks. 




\begin{figure*}[h]
\vspace{-5pt}
	\centering
	\subfigure[Coarse-to-fine on feature pyramid(s).]{\includegraphics[height=0.225\textwidth]{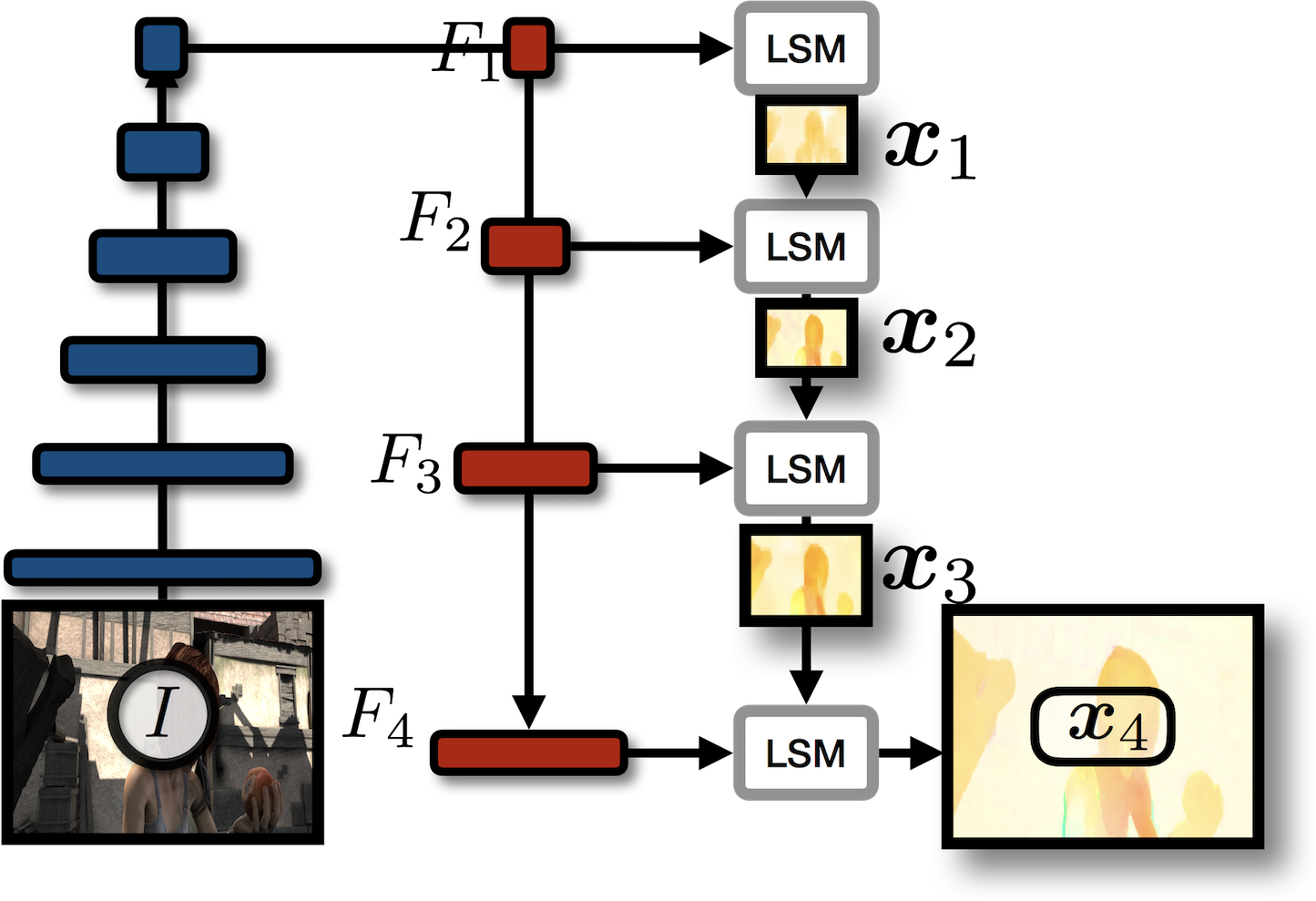}\label{fig:pyramid}}
	\subfigure[A single iteration of the learning subspace minimization.]{\includegraphics[height=0.225\textwidth]{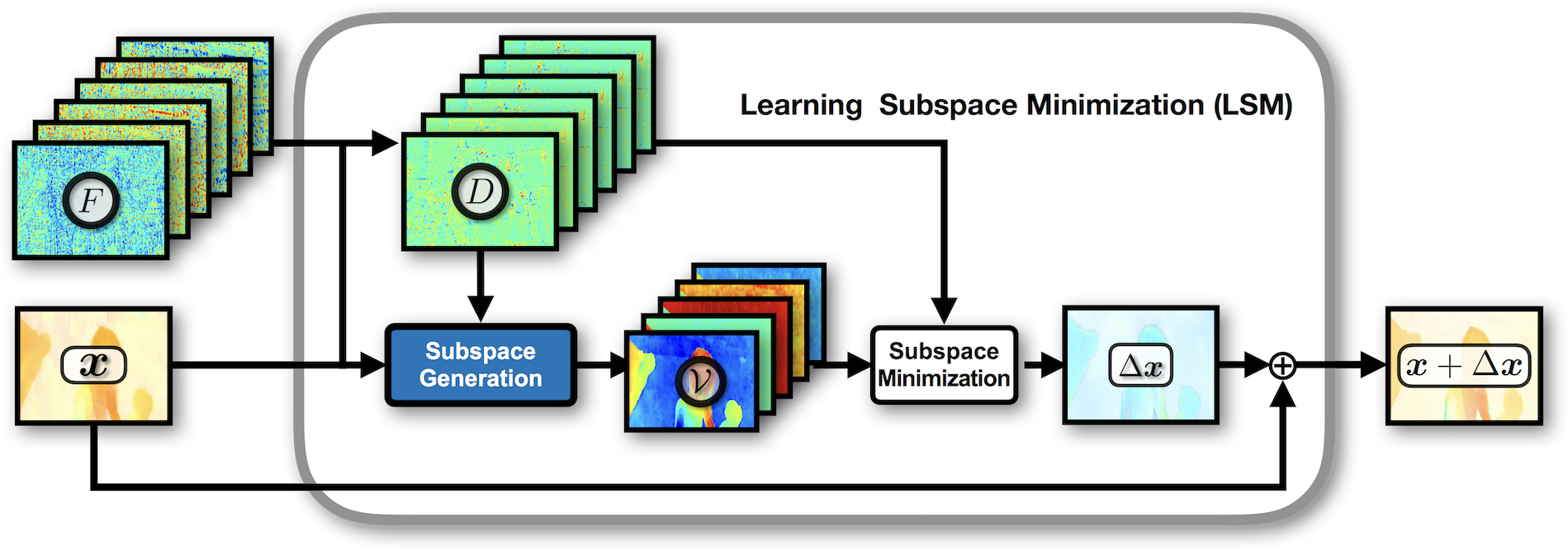}\label{fig:singleiter}}
	\vspace{-5pt}
	\caption{Overview of our learning subspace minimization framework.}
	\vspace{-5pt}
	\label{fig:overview} 
\end{figure*}

\section{Learning Subspace Minimization}
\subsection{Overview}
As illustrated in~\figref{pyramid}, we first build a feature pyramid $\mathcal F$ for~\emph{each} image $I$ from a set, where the number of images in a set depends on the task,~\eg the interactive segmentation is defined on a single image, the stereo matching and the optical flow are defined on two images, and the video segmentation processes three or more images. The output of the pyramid $\mathcal F$ are feature maps in four levels $\{F^{1},F^2,F^3,F^{4}\}$ with strides $\{32,16,8,4\}$ and channels $\{512,256,128,64\}$ respectively, which are constructed by similar strategy as FPN~\citep{FPN} but using DRN-22~\citep{DRN} as the backbone network. 

At each pyramid level, we define the data term $D(\vx)$ of a task on CNN features (\secref{app}) and solve~\equref{2}.~$D(\vx)$ is approximated using the second-order Taylor expansion at the intermediate solution $\vx$ and yields the following quadratic minimization problem:
\vspace{-5pt}
\begin{equation}
\min_{\Delta\vx}\frac{1}{2}\Delta \vx^{\top}\mD\Delta\vx+\vd^{\top}\Delta \vx,
\vspace{-5pt}
\label{equ:iter0}
\end{equation}
where $\mD$ is the matrix that contains the (approximated) second-order derivatives $\frac{\partial^{2}D}{\vx^{2}}$ of the data term, $\vd$ is the vector that contains the first-order derivatives $\frac{\partial D}{\partial\vx}$, and $\Delta \vx$ is the desired incremental solution. The structure of $\mD$ is task dependent: it is a diagonal matrix for one-dimensional tasks or block diagonal for multi-dimensional tasks.  

To maintain the subspace constraint of~\equref{2}, we represent the incremental solution $\Delta \vx$ as a linear combination of a set of underlying basis vectors,~\ie $\Delta \vx=c_1\vv_{1}+c_2\vv_{2}\cdots+c_K\vv_{K}$, and then solve the combination coefficients $\vc=[c_1,c_2\cdots c_K]^{\top}$ as:
\vspace{-5pt}
\begin{equation}
\begin{aligned}
\min_{\vc}\frac{1}{2}\vc^{\top}(\mV^{\top}\mD\mV)\vc+(\vd^{\top}\mV)\vc,
\end{aligned}
\vspace{-5pt}
\label{equ:lsm}
\end{equation}
where $\mV$ is a dense matrix, and its columns correspond to the $K$ basis vectors from $\mathcal{V}$. As shown in~\figref{singleiter}, we generate this $\mathcal{V}$ from the image and the minimization context information (\secref{lsm}), solve minimization with subspace constraint (\secref{minimize}), and move to the next pyramid level after updating the intermediate solution as $\vx\leftarrow\vx+\Delta\vx$. 

This formulation is easy and efficient to implement because multiplying the dense matrix $\mV$ with the (block) diagonal matrix $\mD$ can be done by column-wise product, yielding a compact $K\text{-by-}K$ linear system, which can be solved using direct solver such as Cholesky decomposition~\citep{MATRIX}, instead of iterative solvers such as conjugate gradient descent~\citep{NUMOPT}. Therefore,~\equref{lsm} is differentiable and supports end-to-end training without unrolling or implicit gradient~\citep{Domke2012}.
\subsection{Subspace Generation}
\label{sec:lsm}
\begin{figure*}[h]
\vspace{-5pt}
\centering
\includegraphics[width=\textwidth]{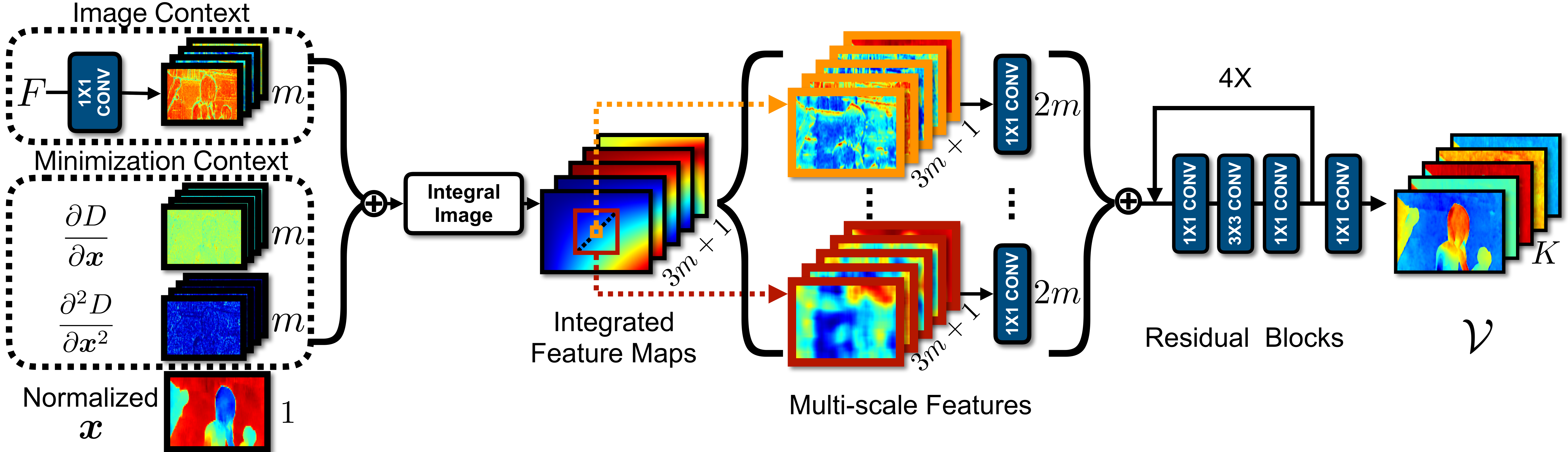}
\vspace{-15pt}
\caption{Subspace generation from the image and the minimization context features. The spatial sizes of all feature maps are the same to the $F$ from a feature pyramid level. Integral image is used for the efficient construction of the multi-scale features.}
\vspace{-5pt}
\label{fig:SL}
\end{figure*}
Before introducing the network that generates $\mathcal{V}$, we first propose two principles for the subspace generation:
\begin{itemize}
\vspace{-5pt}
\item First, \emph{the image context matters.} Standalone data terms are often insufficient for low-level vision tasks as introduced in~\secref{intro}, because these tasks are usually ill-posed~\citep{ILLPOSE,TAREG}. So it is necessary to consider the image context information to generate the subspace $\mathcal{V}$, which enforces each basis vector $\vv_{k}$ to be spatially smooth except for discontinuities at object boundaries.
\vspace{-5pt}
\item Second, \emph{the minimization context matters}. The objective function (data term) is minimized iteratively. At each iteration, the intermediate solution $\vx$ is at a different location on the objective function landscape, and the local curvature of the objective function decides the direction and magnitude of the desired incremental solution $\Delta \vx$ for the minimization. So it is also necessary to incorporate the minimization context into the subspace generation, which learns to narrow the gap between the estimated solution and the ground truth.
\end{itemize}
\vspace{-5pt}
Following these two principles, we learn to generate the subspace $\mathcal{V}$ as illustrated in~\figref{SL}:
\begin{itemize}
\vspace{-5pt}
\item First, we compute a $m$-channel image context from the original $c$-channel feature map $F$ by $1\times1$ convolution, where $m=c/8$ and is $\{64,32,16,8\}$ at the corresponding pyramid level. This step reduces the computation complexity for the following up procedures and balances the impact between the image context and the minimization context.
\vspace{-5pt}
\item Second, we compute a $2m$-channel minimization context. Specifically, we split the $c$-channel feature map(s) into $m$ groups. Within each group, we evaluate the data term $D(x)$ with the associated feature maps, compute the first-order derivative $\frac{\partial D}{\partial\vx}$ and the second-order derivatives $\frac{\partial^{2}D}{\vx^{2}}$, which approximate the objective landscape neighborhood. We concatenate these derivatives to form a $2m$-channel minimization context features.
\vspace{-5pt}
\item In the next, we normalize the intermediate solution $\vx$ with its mean and variance, and concatenate the normalized $\vx$, the image context, and the minimization context to form a $(3m+1)$-channel input features for subspace generation. To aggregate the context information in multi-scale,
we average pool the context features in $4$ different kernel sizes without stride, which maintains the spatial size of a feature map. Specifically, we first compute the integral images~\citep{HARR,GF} of the context features and then average neighboring features at each pixel coordinate, which gives better efficiency.
\vspace{-5pt}
\item Finally, we apply a $1\times1$ convolution to project a feature map to $2m$-channel at each scale individually and concatenate them to get the $8m$-channel multi-scale features. Therefore, we can generate the $K$-dimensional subspace $\mathcal{V}$ from the multi-scale features via four residual blocks~\citep{RESNET} followed by a $1\times1$ convolution, and $K$ is $\{2,4,8,16\}$ at the corresponding pyramid level.
\end{itemize}
\subsection{Subspace Minimization}
\label{sec:minimize}
\begin{figure}[h]
\vspace{-20pt}
	\centering
	\subfigure[$\vx\leftarrow\vx+\mV\vc$]{\includegraphics[height=0.17\textwidth,trim = 0cm 0cm 11.6cm 0cm,clip]{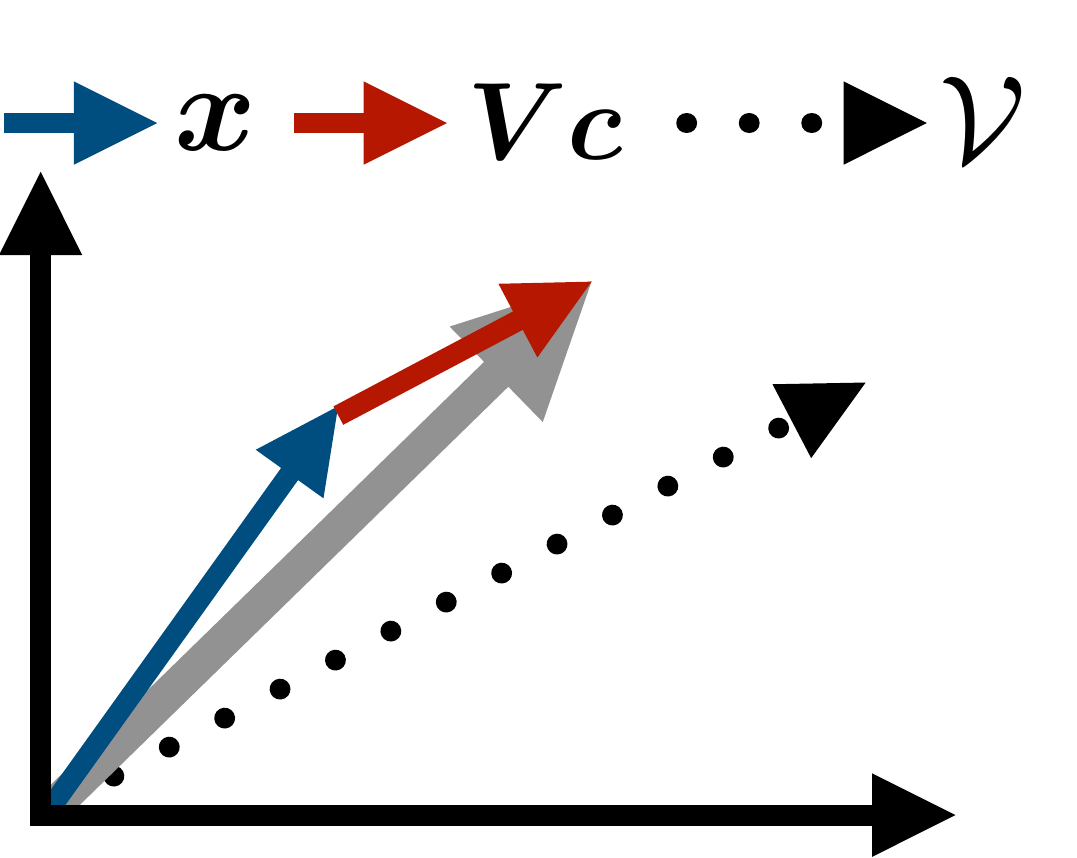}\label{fig:project_off}}
	\subfigure[$\vx\leftarrow\vx+\vr+\mV\vc$]{\includegraphics[height=0.17\textwidth,trim = 11.0cm 0cm 0cm 0cm,clip]{figures/project.pdf}\label{fig:project_on}}
	\vspace{-5pt}
	\caption{A 2D example where (a): the subspace constraint is violated, \ie the 2D vector $\vx+\mV\vc$ is not on the 1D subspace $\mathcal V$, and (b): the subspace constraint is maintained, \ie $\vx+\vr+\mV\vc$ is on $\mathcal V$, by considering the residual $\vr$ between $\vx$ and its projection on $\mathcal V$.}
	\vspace{-5pt}
	\label{fig:project} 
\end{figure}
After the subspace $\mathcal V$ is generated, we can solve~\equref{lsm} directly as $\vc=-(\mV^{\top}\mD\mV)^{-1}\mV^{\top}\vd$ because $\mV^{\top}\mD\mV$ is positive-definite by definition, and update the current intermediate solution as $\vx\leftarrow\vx+\mV\vc$. However, it will violate the subspace constraint as shown in~\figref{project_off}, because the subspace $\mathcal{V}$ is generated progressively, i.e, the current solution $\vx$ belongs to the subspace from last iteration but is not guaranteed to be on the newly generated $\mathcal{V}$, so is $\vx+\mV\vc$.
To address this issue, we propose to project $\vx$ onto the current subspace $\mathcal V$ and reformulate~\equref{lsm} as follows:
\vspace{-5pt}
\begin{itemize}
\item Denoting $\mP=\mV(\mV^{\top}\mV)^{-1}\mV^{\top}$ is the projection matrix that projects an arbitrary vector onto the the subspace $\mathcal V$, we can compute its projection onto $\mathcal V$ as $\vx^{\prime}=\mP\vx$, and the residual vector from $\vx$ to $\vx^{\prime}$ is $\vr=(\mP-\mI)\vx$.
\vspace{-5pt}
\item Theoretically, we can reevaluate $\mD$ and $\vd$ respect to $\vx^{\prime}$ and solve~\equref{lsm}, but it requires extra computation. So we reparameterize the incremental solution $\Delta\vx$ as $\vr+\mV\vc$ and transform~\equref{lsm} into:
\vspace{-5pt}
\begin{equation}
\min_{\vc}\frac{1}{2}(\vr+\mV\vc)^{\top}\mD(\vr+\mV\vc)+\vd^{\top}(\vr+\mV\vc),
\vspace{-5pt}
\label{equ:min_proj}
\end{equation} 
\vspace{-5pt}
where we can compute $\vc$ as
\begin{equation}
\vspace{-5pt}
\vc=-(\mV^{\top}\mD\mV)^{-1}\mV^{\top}(\vd+\mD\vr)
\vspace{-2pt}
\end{equation}
without recomputing $\mD$ and $\vd$, and update $\vx$ as $\vx+\vr+\mV\vc$ as shown in~\figref{project_on}.
\end{itemize}

\subsection{Applications}
\label{sec:app}
We now show how the proposed LSM framework unifies various low-level vision tasks. We implement four tasks for demonstration, and only introduce the data term for each task. For~\emph{all} tasks, we initialize $\vx$ as a zero vector. According to the difference of data term formulation, these tasks are divided to two categories.~\label{sec:app}

In the first category, we introduce two binary image labeling tasks: interactive segmentation and video segmentation, both of which share the same formulation as:
\vspace{-5pt}
\begin{equation}
D(\vx)=\sum_{\vp}\alpha_\vp\|\tau(\vx_\vp)-1\|_{2}^{2}+\beta_\vp\|\tau(\vx_\vp)+1\|_{2}^{2},
\label{equ:edge}
\vspace{-5pt}
\end{equation}
where $\vp=[x,y]$ is a pixel coordinate, $\tau(\cdot)$ is an activation function to relax and constrain the binary label and constrain $\tau(\vx_\vp)$ between $(-1,+1)$, while $\alpha_\vp$ and $\beta_\vp$ are the probabilities that $\tau(\vx_{\vp})=+1 \text{ or}-1$.

\begin{itemize}
\vspace{-5pt}
\item For \textbf{interactive segmentation}, $\tau(\vx_\vp)$ indicates whether a pixel ${\vp}$ is on the foreground object ($+1$) or background scene ($-1$), and the corresponding probabilities ${\alpha_\vp}$ and $\beta_\vp$ are estimated as nonparametric probabilities~\citep{Wu_2018_CVPR} from the foreground scribble points and the background scribble points respectively.
\vspace{-5pt}
\item For \textbf{video segmentation}, $\tau(\vx_\vp)$ indicates whether a pixel ${\vp}$ belongs to an previously labeled foreground object ($+1$) or not ($-1$), and ${\alpha_\vp}$ and $\beta_\vp$ are the corresponding average probabilities estimated from $\vp$'s correlation with its foreground and background neighbors in previous labeled frames respectively.
\vspace{-5pt}
\end{itemize}
In the second category, we introduce two dense correspondence estimation tasks on two images: stereo matching and optical flow,  both of which can be formulated as:
\vspace{-5pt}
\begin{equation}
D(\vx)=\sum_{\vp}\|F_{S}(\vp+\vx_{\vp})-F_{T}(\vp)\|_{2}^{2},
\label{equ:corr}
\vspace{-5pt}
\end{equation}
where $\vp=[x,y]$ is the pixel coordinate in the target (template) image $T$, and $\vx_{\vp}$ is the warping vector that warps ${\vp}$ to $\vp+\vx_{\vp}$ in the source image $S$. Similar to the brightness constancy assumption for image channels~\citep{HS},~\equref{corr} assumes that the warped feature channels $F$ will also be consistent.
\begin{itemize}
\vspace{-5pt}
\item For~\textbf{stereo matching}, $S$ and $T$ are two images viewing the same scene.
Therefore, $\vx_{\vp}=[u,0]$ only contains horizontal displacement and warps $\vp$ to $[x+u,y]$ in the target image $T$.
\vspace{-5pt}
\item For~\textbf{optical flow}, $S$ and $T$ are two neighboring video frames. Therefore, $\vx_{\vp}=[u,v]$ is the 2D motion vector that warps $\vp$ to $[x+u,y+v]$ in the $S$. Since optical flow is a two-dimensional labeling problem compared with stereo matching (one-dimensional, \ie $\vx_{\vp}$ is a scalar) and the two image labeling tasks, we apply Cramer's rule~\citep{Higham} to unify the network structures and parameters of optical flow with others. Please refer to the~\emph{supplementary} for more implementation details.
\vspace{-5pt}
\end{itemize}
\section{Experiments}

\subsection{Implementation Details}
\label{sec:training}
\paragraph{Training Loss} Loss design is beyond the scope of this paper, so we use existing losses for all tasks. For~\emph{interactive segmentation} and~\emph{video segmentation}, we use the Intersection of Union (IoU) loss from~\cite{IOU}. For~\emph{stereo matching} and~\emph{optical flow} we use the end-point-error (EPE) loss as in DispNet~\citep{SYNDATA} and FlowNet~\citep{FLOWNET}. Since our solution is estimated coarse-to-fine, we downsample the ground-truth to multiple scales and sum the loss over all scales as in~\citep{PWCNET}.

\paragraph{Hyperparamters}
We use AdamW optimizer~\citep{ADAMW} with the default settings where $\beta_{1}=0.9$, $\beta_{2}=0.999$. The learning rate is initialized as $3\times10^{-4}$ and reduced during training using cosine decay~\citep{SGDR} without warm restarts. This set of hyperparameters are fixed for all experiments. 

\paragraph{Dataset}
For \emph{interactive segmentation}, we use the PASCAL VOC Semantic Boundaries Dataset~\citep{BharathICCV2011} for training and the VGG interactive segmentation dataset~\citep{Gulshan10} for testing, and the overlapped 99 images are excluded from the training set. 
For \emph{video segmentation}, we use the DAVIS-2017 dataset~\citep{2017} for training and the DAVIS-2016~\citep{2016} for testing. 
For~\emph{stereo matching}, we use the training/testing split of FlyingThings3D~\citep{SYNDATA} from~\citep{CRL}, and for~\emph{optical flow} we use FlyingThings3D for training and Sintel~\citep{SINTEL} for testing.

\subsection{Comparison with State-of-the-art}
\label{sec:multi_task}
Our framework can be applied to a low-level vision task as long as its first-order and second-order differentiable data term can be formulated.
So we first test the multi-task capability of our network. Note that the \emph{whole} network structure and \emph{all} parameters are shared for all tasks, while previous works~\citep{MULTI1,MULTI2,UBER} \emph{only} share the backbone and use different decoders/heads to handle different tasks. 

We train our model on all four tasks jointly using a workstation with four TITAN-Xp GPUs. For implementation simplicity, we deploy one task on each GPU and update the network parameters on CPU. The batch size are 12 for interactive segmentation, 6 for video segmentation, and 4 for stereo matching and optical flow. The training runs for 143.2K iterations. To make a fair comparison with other state-of-the-art single-task methods, we also train each task individually and denote the corresponding result as `Single', while the results of joint training are denoted as `Joint'.

\begin{figure}[h]
\vspace{-5pt}
	\centering
	\vspace{-5pt}\subfigure[Average plots required to make IoU>0.85.]{\includegraphics[width=\linewidth,page=1]{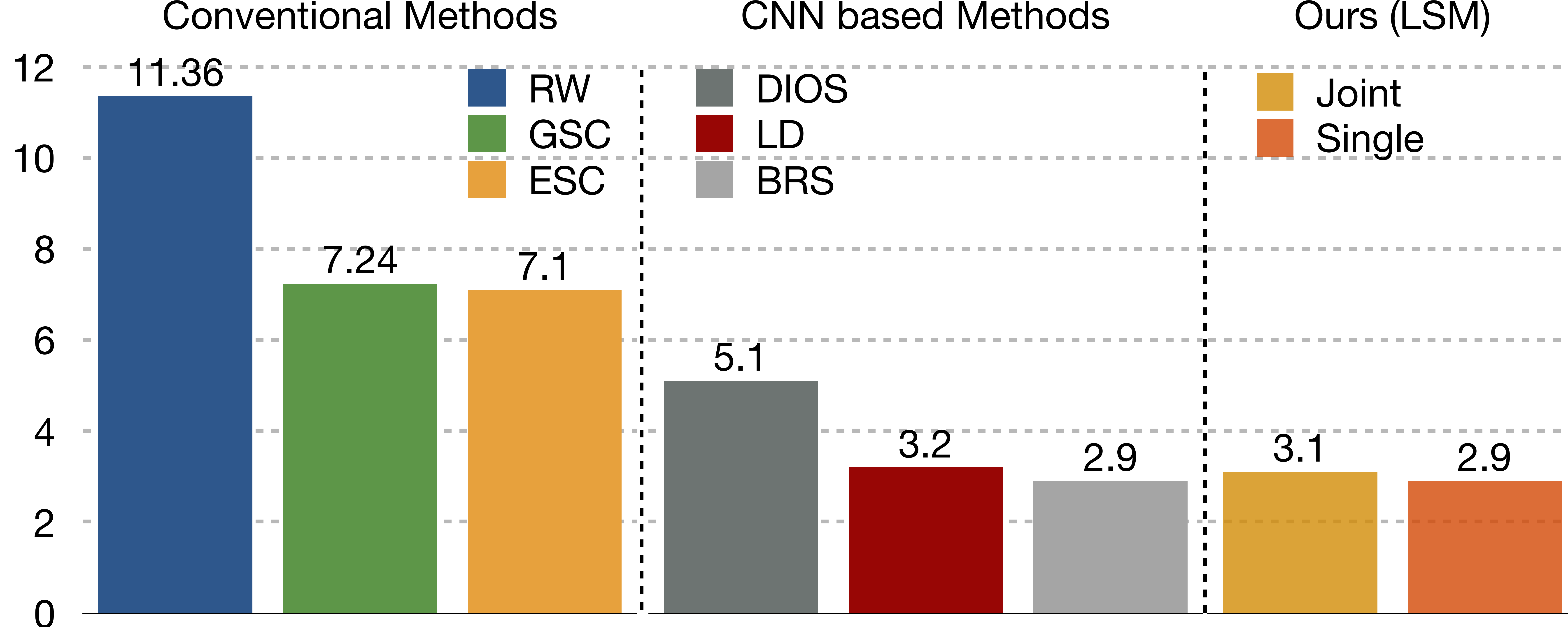}\label{fig:seg1}}
	\vspace{-7pt}\subfigure[Our result is superior than LD~\citep{LATENT} when user interact only once.]{\includegraphics[width=\linewidth]{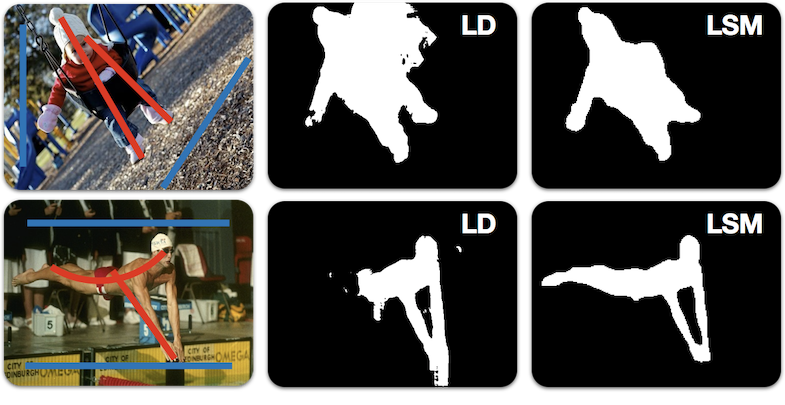}\label{fig:seg2}}
	\vspace{-5pt}
	\caption{Interactive image segmentation result on VGG interactive segmentation benchmark.}
	\vspace{-10pt}
	\label{fig:seg}
\end{figure}

\paragraph{Interactive Image Segmentation}
For interactive segmentation, we compare our LSM framework to several conventional methods including ESC and GSC by~\cite{Gulshan10}, and Random walk~\citep{RANDOMWALK}, as well as recent CNN based methods Deep Object Selection (DIOS)~\citep{DIOS}, Latent Diversity (LD)~\citep{LATENT} and Backpropagation Refinement (BRS)~\citep{BRS}. We randomly sample a few points from the scribbles as inputs for the CNN based methods, since they only supports clicks. We evaluate all methods by the plots required to make IoU greater than 0.85. As shown in~\figref{seg1}, our method achieves better results among both recent CNN based methods and the conventional ones.

We also compare with the LD qualitatively when user only interact once. We also subsample scribbles and successively send annotations to LD for a fair comparison.~\figref{seg2} shows that our results are superior than Latent Diversity~\citep{LATENT}. It is because the existing CNN based methods only supports spatial distance maps as inputs, which are less precise than scribbles. While our LSM supports scribbles by feature distribution estimation and~\equref{edge}. 

\begin{figure}[h]
	\centering
	\vspace{-5pt}\subfigure[Average IoU for video segmentation.]{\includegraphics[width=\linewidth,page=4]{figures/compare.pdf}\label{fig:vidseg1}}
	\vspace{-7pt}\subfigure[Our result is qualitatively comparable to FEELVOS~\citep{FEELVOS}.]{\includegraphics[width=\linewidth]{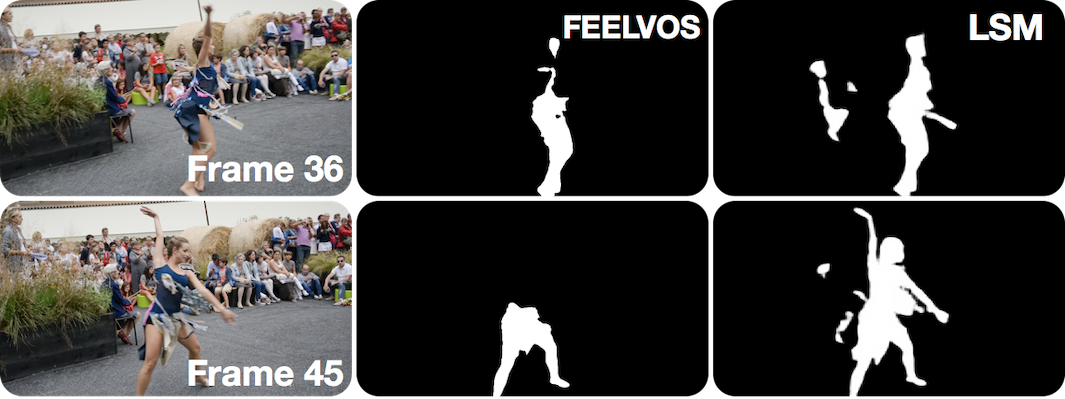}\label{fig:vidseg2}}
	\vspace{-5pt}
	\caption{Video segmentation result on DAVIS 2016.}
	\label{fig:vidseg}
	\vspace{-5pt}
\end{figure}

\paragraph{Video Segmentation}
For video segmentation, we compare our LSM framework to several conventional minimization based methods including BVS~\citep{BVS}, OFL~\citep{OFL} and DAC~\citep{DAC}, as well as recent CNN based methods that do not require fine-tuning for a fair comparison, including MaskRNN~\citep{MASKRNN}, VideoMatch~\citep{VIDEOMATCH} and FEELVOS~\citep{FEELVOS}.~\figref{vidseg1} shows that our LSM performs better than conventional methods and comparable to CNN based methods. 
We also show a qualitative comparison to FEELVOS on the challenging dance-twirl sequence. As shown in~\figref{vidseg2}, our LSM generates more false positive regions than FEELVOS~\citep{FEELVOS} because the skin and the cloth colors of the dancer and the audiences are similar, but ours is able to track the dancer consistently while FEELVOS lost the dancer's torso during twirl.

\paragraph{Stereo Matching}
For stereo matching, we compare our LSM framework with several conventional methods including SGM~\citep{DSPSGM}, SPS~\citep{Yamaguchi14} and MC-CNN~\citep{MCNET} which uses CNN features~\emph{only} for data term evaluation in a MRF, as well as some fully CNN based methods including DispNet~\citep{SYNDATA}, CRL~\citep{CRL}, PSMNet~\citep{PSMNET} and GANet~\citep{GANET}. 

\begin{figure}[ht]
\vspace{-5pt}
	\centering
	\vspace{-5pt}\subfigure[Average End-Point-Error for disparity.]{\includegraphics[width=\linewidth,page=2]{figures/compare.pdf}\label{fig:things_epe_disp}}
	\vspace{-7pt}\subfigure[Our LSM supports both left-to-right and right-to-left stereo matching while most of fully CNN based methods only support left-to-right.]{\includegraphics[width=\linewidth]{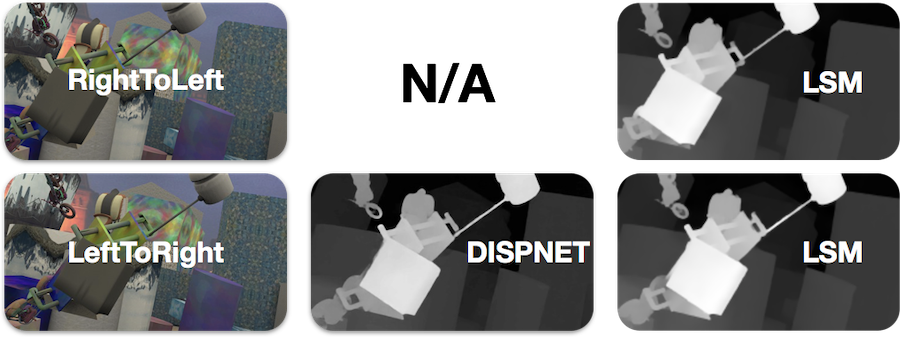}\label{fig:l2r}}
	\vspace{-5pt}
	\caption{Stereo matching results on FlyingThings 3D.}
	\vspace{-5pt}
	\label{fig:things}
\end{figure}

When compared with other CNN based methods, our LSM is comparable for joint training and better for single-task training as shown in~\figref{things_epe_disp}. As shown in~\figref{l2r}, we are able to estimate both the left-to-right and the right-to-left disparities in the same accuracy because we do not assume the direction or the range of the disparity in~\equref{corr}. Altough~\citep{PDS} has been proposed to achieve range flexibility, the fully CNN based methods still only deal with single directional pairs because of the cost-volume. 

\paragraph{Optical Flow}
For optical flow, we compare our LSM framework with conventional methods including LDOF~\citep{LARGFLO}, EpicFlow~\citep{EPIC} and PCA-Layers~\citep{PCAFLO} which also adopts a basis representation but the basis is static and specifically learned for optical flow using PCA~\citep{PCABook}, as well as CNN based methods including LiteFlowNet~\citep{LITEFLOW}, PWC-Net~\citep{PWCNET}, and FlowNet2-CSS~\citep{FLOWNET2} which is a stack of three FlowNets. 

\begin{figure}[h]
\vspace{-5pt}
	\centering
	\vspace{-5pt}\subfigure[Average End-Point-Error]{\includegraphics[width=\linewidth,page=3]{figures/compare.pdf}\label{fig:flow1}}
	\vspace{-7pt}\subfigure[Our optical flow is comparable to PWC-Net.]{\includegraphics[width=\linewidth]{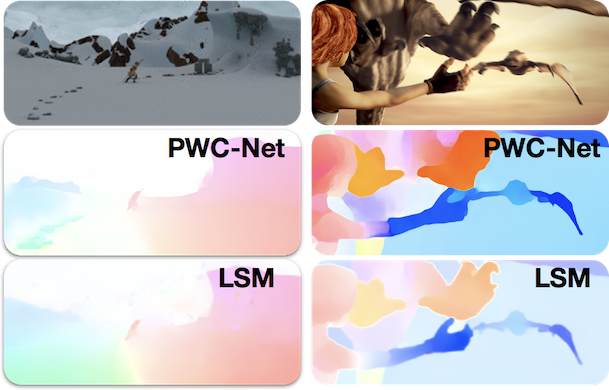}\label{fig:flow2}}
	\vspace{-5pt}
	\caption{Flow trained on Flythings3D and tested on Sintel.}
	\vspace{-5pt}
	\label{fig:flow}
\end{figure}

As shown in~\figref{flow1}, our result are comparable to LiteFlowNet~\citep{LITEFLOW} and PWC-Net~\citep{PWCNET} without refinement sub-net. FlowNet2 is more accurate by stacking networks, which is less efficient, more difficult to train, and increases the model size dramatically. Comparing with FlowNet2, our method is $12\times$ smaller in model size, $4\times$ faster in inference, and $32\times$ less in training time. Our LSM is better than LDOF~\citep{LARGFLO} and PCA-Layers~\citep{PCAFLO}, but less accurate than EpicFlow~\citep{EPIC}. However, conventional method are usually based on variational approaches and take 5-20 seconds to run, while our LSM takes only 25ms.
\vspace{-10pt}
\subsection{Zero-shot Task Generalization}
Our LSM framework even generalizes the learned network to unseen tasks. It is different from the zero-shot task transfer~\citep{ZEROTANSFER}, where the network parameters are interpolated from existing tasks, and the interpolation coefficients is defined by a correlation matrix during training. In contrast, we fix the learned parameters and do not require any extra information between tasks. To demonstrate this capability, we train the network on three tasks with the same settings as the joint multi-task training, and leave one out for testing.
\vspace{-10pt}
\begin{figure}[h]
	\centering
	\includegraphics[width=\linewidth]{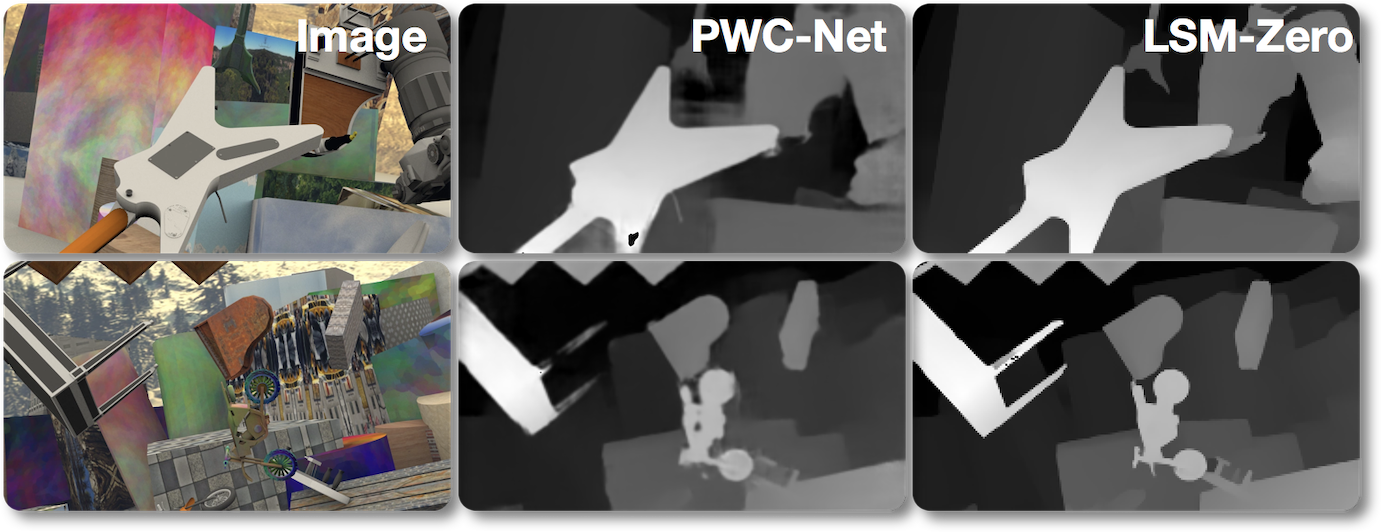}
	\vspace{-15pt}
	\caption{Our zero-shot generalized LSM model performs better than PWC-Net for stereo matching.}
	\vspace{-5pt}
	\label{fig:stereo0}
\end{figure}

\paragraph{Stereo Matching} 
The first task left out for testing is stereo matching. Since none of existing CNN based method supports this test, we approximate it by estimating optical flow using PWC-Net~\citep{PWCNET} on stereo image pairs, and only consider the EPE on the horizontal direction. 
The average EPE is 2.47 for our LSM model learned on the other three tasks and tested on stereo matching, which is superior than the 5.29 EPE of PWC-Net as shown in~\figref{stereo0}. Note that our LSM consistently performs better than conventional methods~\citep{DSPSGM,Yamaguchi14,MCNET}, while PWC-Net is worse than SGM~\citep{DSPSGM}.
\vspace{-10pt}
\begin{figure}[h]
	\centering
	\includegraphics[width=\linewidth]{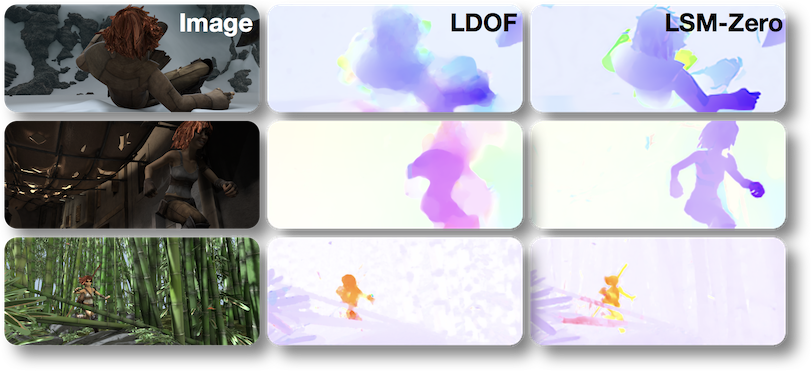}
	\vspace{-15pt}
	\caption{Our zero-shot generalized LSM model performs better than LDOF for optical flow.}
	\vspace{-5pt}
	\label{fig:flow0}
\end{figure}

\paragraph{Optical Flow} For optical flow, none of CNN based method supports this zero-shot test, and the average EPE is 4.6 for our LSM model learned on the other three tasks, which is better than LDOF~\citep{LARGFLO}. However, LDOF requires computationally expensive dense HOG~\citep{HOG} feature matching as external input, while our LSM estimates the optical flow efficiently only by minimizing the feature-metric alignment error in~\equref{corr}.~\figref{flow0} shows that our zero-shot optical flow maintains the object-aware discontinuities, which indicates that the subspace generator learned from the other three tasks is general, while LDOF generates over-smoothed results because it uses the $L2$ smoothness regularization term.
\vspace{-10pt}
\begin{figure}[h]
	\centering
	\includegraphics[width=\linewidth]{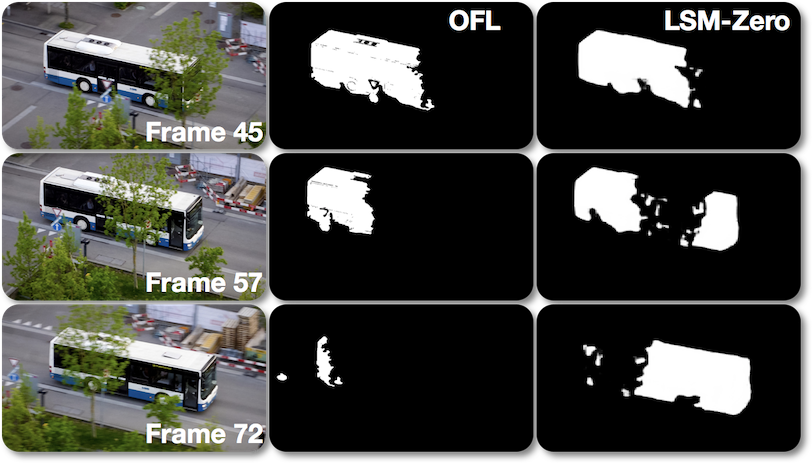}
	\vspace{-15pt}
	\caption{Our zero-shot generalized LSM model is more robust to occlusion than OFL~\citep{OFL} for video segmentation.}
	\vspace{-5pt}
	\label{fig:vidseg0}
\end{figure}

\paragraph{Video Segmentation} 
The third task left out for testing is video segmentation. The average IoU is 0.682 for our LSM model learned on the other tasks and tested on video segmentation, which is comparable to conventional methods such as OFL~\citep{OFL}. However, as shown in~\figref{vidseg0}, our method is more robust to partial occlusions, while OFL lost tracking of the bus when partially occluded by trees. Please refer to the~\emph{supplementary} for the zero-shot generalization test on interactive image segmentation due to the page limit.
\subsection{Ablation Studies}
\vspace{-10pt}
\begin{figure}[ht]
	\centering
    \includegraphics[width=\linewidth,height=0.6\linewidth]{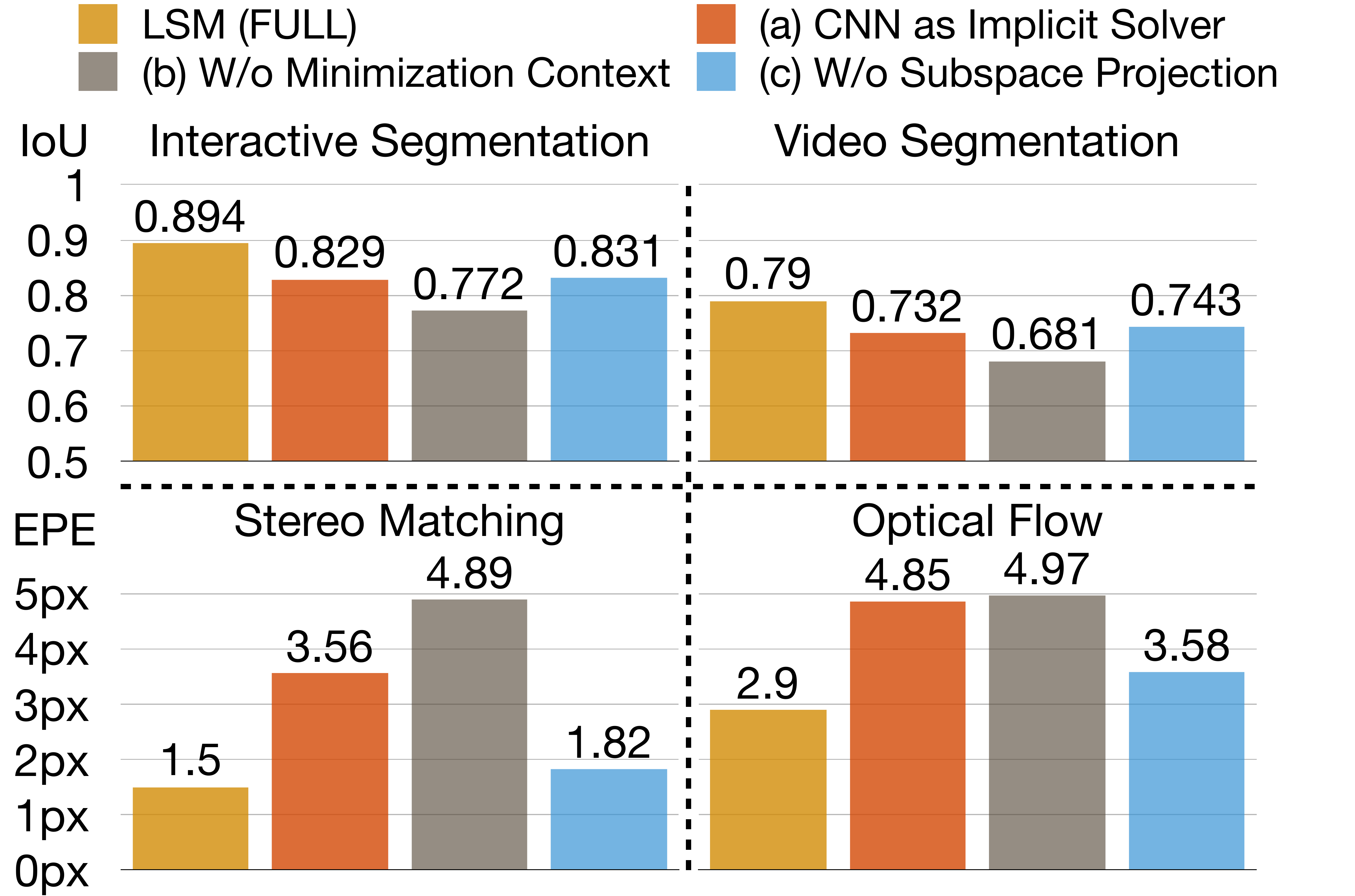}
    \vspace{-20pt}
	\caption{Ablation studies on (a) solving~\equref{iter0} using CNNs as implicit solver, (b) generating subspace $\mathcal V$ without minimization context, and (c) minimization without subspace projection, \ie directly use~\equref{lsm}.}
	\vspace{-5pt}
	\label{fig:wo}
\end{figure}

\paragraph{CNN as Implicit Solver} The first question is whether the explicit minimization is necessary, i.e can we use CNN as an implicit solver and predict the solution directly from the image and the minimization context features? To answer this question, we keep the same network structure except the last convolution layer of the subspace generators, \ie the output of the subspace generator is reduced to one-channel and directly serves as the solution~$\vx$. Then the subspace generator becomes an implicit minimization solver, and the modified network is trained with the same training protocol.

As shown in~\figref{wo}, without minimization, the interactive segmentation and the video segmentation's get lower IoU while the stereo matching and the optical flow get higher EPE, which indicates the explicit minimization is preferred than learning to minimize via CNNs for our LSM framework.

\paragraph{Without Minimization Context} The second question is whether it is necessary to incorporate the minimization context into the subspace generation, i.e can we predict the subspace solely from the image features as in Code-SLAM~\citep{CODESLAM} and BA-Net~\citep{BA}? To answer this question, we predict the subspace without minimization context and keep the same network structure except the first several convolution layers after the multi-scale context features. The modified network is also trained with the same training protocol in~\secref{training}.

As shown in~\figref{wo}, all the four tasks performs significantly worse without the minimization context, which indicates the minimization context is necessary for subspace generation. It is difficult to learn an unified subspace generator solely from image context, because different tasks requires different subspace even on the same image.

\paragraph{Without Subspace Projection}
Finally, we evaluate the effectiveness of the subspace projection proposed in~\secref{minimize}, \ie minimizing~\equref{lsm} instead of ~\equref{min_proj}. We also train the modified network for a fair comparison.

As shown in~\figref{wo}, the network without the subspace projection performs worse than the original full pipeline, which indicates that maintaining the subspace constraint via projection is necessary not only in theory but also in practice for better performance. It is because, with the subspace projection, the predicted subspace $\mathcal{V}$ is learned to be consistently towards the ground truth solution. In contrast, learning without projection violates the subspace constraint, and make the minimization less constrained and training more difficult.

\subsection{Visualization of Generated Subspaces}
\vspace{-15pt}
\begin{figure}[h]
\includegraphics[width=\linewidth]{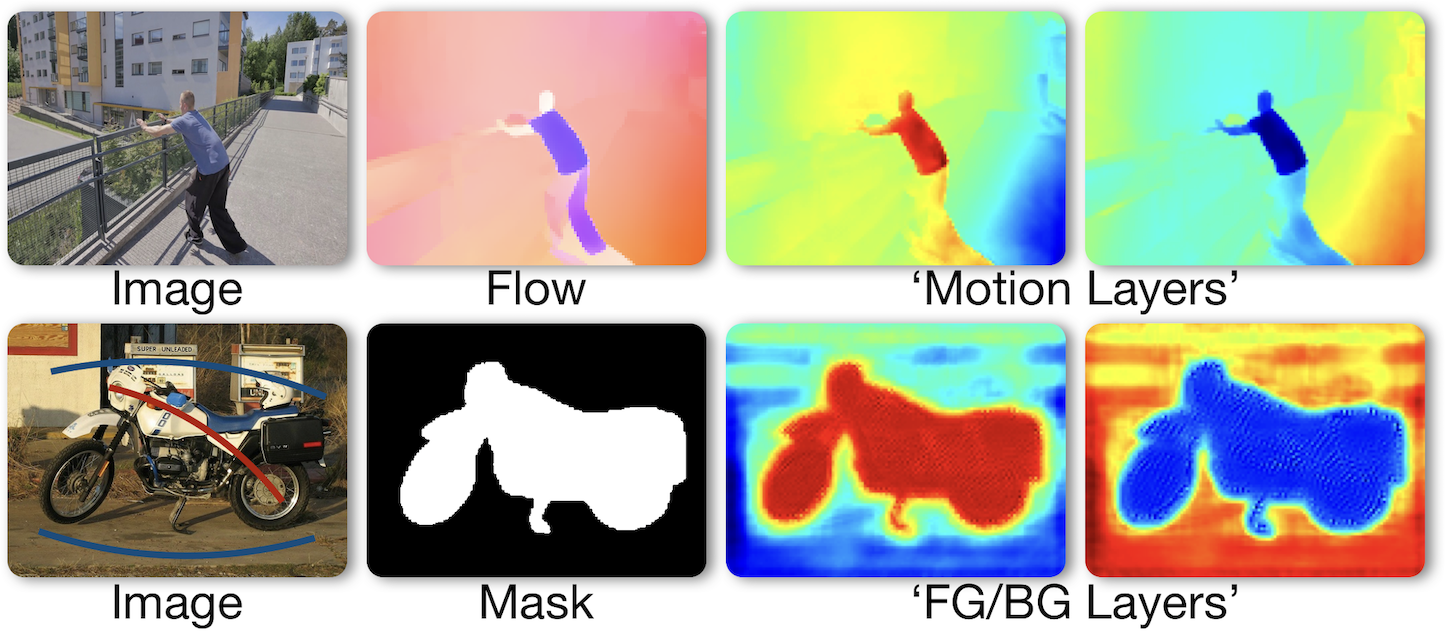}
\vspace{-20pt}
\caption{Visualization of generated subspaces for optical flow and interactive segmentation.}
\vspace{-5pt}
\end{figure}
As introduced in~\secref{intro}, the intuition of using a subspace constraint is the solution of a low-level task is usually composed of several layers. To verify whether the generated subspaces satisfy this intuition, we visualize some basis vectors as heat maps for the optical flow and the interactive segmentation tasks. As we can see, the basis vectors are consistent with the motion layers for optical flow and the foreground/background layers for segmentation, which also indicates that our subspace generation network captures the intrinsic characteristics of each task.
\section{Conclusions}
We propose the learning subspace minimization (LSM) framework to address low-level vision problems that can be formulated as an energy minimization of a data term and a regularization term.
We learn convolution neural networks to generate a content-aware subspace constraint to replace the regularization term which is often heuristic and hinders performance.
At the same time, we exploit the data term and minimize it to solve a low-level task, because the data term is often derived from the first principle  of a task and captures the underlying nature of a problem.
This approach nicely combines domain knowledge (\ie minimizing data terms derived from first principles) and the expressive power of CNNs (\ie learning to predict content-aware subspace constraint).
Our LSM framework supports joint multi-task learning with completely shared parameters and also generate state-of-the-art results with much smaller network and faster computation. It even enables zero-shot task generalization, where a trained network can be generalized to unseen tasks. This capability demonstrates our LSM framework can be applied to a wide range of computer vision tasks.

{\small
\bibliographystyle{plainnat}
\bibliography{iclr2020_conference}
}
\appendix
\section{Derivatives of Various Data Terms $D(\vx)$}
\label{sec:a_app}
In Sec~3.4, we introduced two categories of tasks. Now, we show the first-order and the (approximated) second-order derivatives of the data terms, which compose the vector $\vd$ and the (block) diagonal matrix $\mD$ at each iteration. 

\paragraph{Binary Image Labeling} Recall that the first category is binary image labeling (interactive segmentation and video segmentation) as:
\begin{equation}
D(\vx)=\sum_{\vp}\alpha_\vp\|\tau(\vx_\vp)-1\|_{2}^{2}+\beta_\vp\|\tau(\vx_\vp)+1\|_{2}^{2},
\label{equ:bin}
\end{equation}
where $\vp=[x,y]^{\top}$ is a pixel coordinate, $\tau$ is an activation function to relax the binary label $\tau(\vx_\vp)$ between $(+1,-1)$, and $\alpha_\vp$ and $\beta_\vp$ are the probabilities that $\tau(\vx_{\vp})=+1 \text{ or}-1$. Therefore, the first-order and the second-order derivatives at an intermediate solution $\vx$ are:
\begin{equation}
\begin{aligned}
\frac{\partial D}{\partial\vx_\vp}=&[(\alpha_{\vp}+\beta_{\vp})\tau(\vx_\vp)+(\beta_{\vp}-\alpha_{\vp})][\frac{\partial\tau(\vx_\vp)}{\partial\vx_\vp}],\\
\frac{\partial^{2} D}{\partial\vx_\vp^{2}}=&(\alpha_{\vp}+\beta_{\vp})[\frac{\partial\tau(\vx_\vp)}{\partial\vx_\vp}]^{2},
\end{aligned}
\end{equation}
where we ignore the scale factor $2$ for simplicity, and $\frac{\partial\tau(\vx_\vp)}{\partial\vx_\vp}$ can be $1-\tau^{2}(\vx_\vp)$ for $tanh$ activation function.

\paragraph{Dense Correspondence Estimation} The second category is the dense correspondence estimation (stereo matching and optical flow) where the data term is:
\begin{equation}
D(\vx)=\sum_{\vp}\|F_{S}(\vp+\vx_{\vp})-F_{T}(\vp)\|_{2}^{2}.
\label{equ:corr}
\end{equation}
For stereo matching, the derivatives are derived as:
\begin{equation}
\begin{aligned}
\frac{\partial D}{\partial\vx_\vp}=&\nabla_{x} F_{S}(\vp+\vx_{\vp})^{\top}[F_{S}(\vp+\vx_{\vp})-F_{T}(\vp)],\\
\frac{\partial^{2} D}{\partial\vx_\vp^{2}}=&\|\nabla_{x} F_{S}(\vp+\vx_{\vp})\|_{2}^{2},
\end{aligned}
\label{equ:stereo0}
\end{equation}
where $\nabla_{x}$ is the gradient operator along the horizontal direction. $\nabla_{x} F_{S}(\vp+\vx_{\vp})$ and $[F_{S}(\vp+\vx_{\vp})-F_{T}(\vp)]$ are vectors, so $\frac{\partial D}{\partial\vx_\vp}$ and $\frac{\partial^{2} D}{\partial\vx_\vp^{2}}$ are scalars, which is also an one-dimensional problem and can be unified with the binary image label tasks with the same network and the parameters. 
\begin{figure*}
\vspace{-5pt}
	\centering
	\includegraphics[width=\textwidth,page=1]{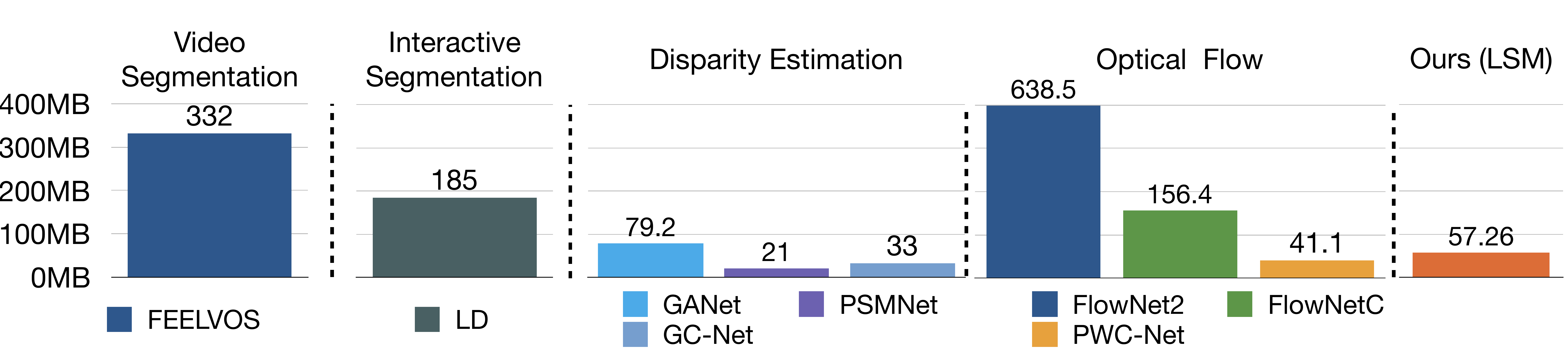}\label{fig:size}
	\vspace{-5pt}
	\caption{Our LSM model handles multiple tasks in a relatively small model.}
	\vspace{-5pt}
	\label{fig:size} 
\end{figure*}
For optical flow, $\vx_{\vp}=[u,v]^{\top}$ is a 2D vector and the derivatives are:
\begin{equation}
\begin{aligned}
\frac{\partial D}{\partial\vx_\vp}=&\nabla F_{S}(\vp+\vx_{\vp})^{\top}[F_{S}(\vp+\vx_{\vp})-F_{T}(\vp)],\\
\frac{\partial^{2} D}{\partial\vx_\vp^{2}}=&\nabla F_{S}(\vp+\vx_{\vp})^{\top}\nabla F_{S}(\vp+\vx_{\vp}),
\end{aligned}
\label{equ:flow0}
\end{equation}
where $\nabla$ is the gradient operator along both the horizontal and vertical direction. Therefore, $\frac{\partial D}{\partial\vx_\vp}$ is a $2\times1$ vector, and $\frac{\partial^{2} D}{\partial\vx_\vp^{2}}$ is a $2\times2$ matrix, which makes unification with other one-dimensional tasks difficult. To address this problem, we apply Cramer’s rule~\citep{Higham} as follows:
\begin{itemize}
\vspace{-5pt}
\item First, we compute the determinant of $\frac{\partial^{2} D}{\partial\vx_\vp^{2}}$ as $det_{\vp}$. 
\vspace{-5pt}
\item Next, we replace the first column of $\frac{\partial^{2} D}{\partial\vx_\vp^{2}}$ with $\frac{\partial D}{\partial\vx_\vp}$, and denote the determinant of the modified matrix as $det^{x}_{\vp}$. Similarly, $det^{y}_{\vp}$ is computed by replacing the second column of $\frac{\partial^{2} D}{\partial\vx_\vp^{2}}$ with $\frac{\partial D}{\partial\vx_\vp}$. 
\vspace{-5pt}
\item Finally, we collect $det^{x}_{\vp}$ and $det_{\vp}$ at all pixel locations as the minimization context, concatenate it with the image context to generate the subspace $\mathcal{V}_{x}$ for the horizontal component of the flow field. Similarly, the $det^{y}_{\vp}$ and the $det_{\vp}$ are collected as the minimization context for the vertical subspace $\mathcal{V}_{y}$. Thus the subspace generation for optical flow is unified with other one-dimensional tasks by generating the subspace for the horizontal and the vertical components of flow individually.
\end{itemize}

\section{Model Efficiency}
Our LSM model is efficient in terms of model size, training time, and inference time, which are contributed by integrating data terms explicitly.

\subsection{Model Size}
We implement our LSM framework with the aforementioned settings, which contains about 15M parameters and costs 57.26 MB in memory. As shown in~\figref{size}, our LSM model maintains a relatively small model size when compared with other CNN based methods. But our LSM model handles multiple tasks within the same parameters while others are designed specifically for single tasks.

\subsection{Training Efficiency}
We train our model with 143.2K iterations for all the experiments, which tasks roughly 20 hours and is relatively faster compared to existing CNN based methods. For example, training FlowNet2~\citep{FLOWNET2} tasks more than 14 days and PWC-Net~\citep{PWCNET} takes 4.8 days. We initialize the backbone DRN-22 from the ImageNet pre-trained model, which also helps the training converges faster~\citep{RETHINK}. 

\subsection{Inference Efficiency}
Our LSM framework is also efficient during inference. Since we unify different tasks into a single network, the inference times for various tasks are roughly the same, which consume about 25ms for $512\times384$ images. The computation is dominated by the feature pyramid construction, the subspace generation and the minimization.

\section{Zero-shot Interactive Segmentation}
\begin{figure}[h]
	\vspace{-5pt}
	\centering
	\includegraphics[width=0.99\linewidth]{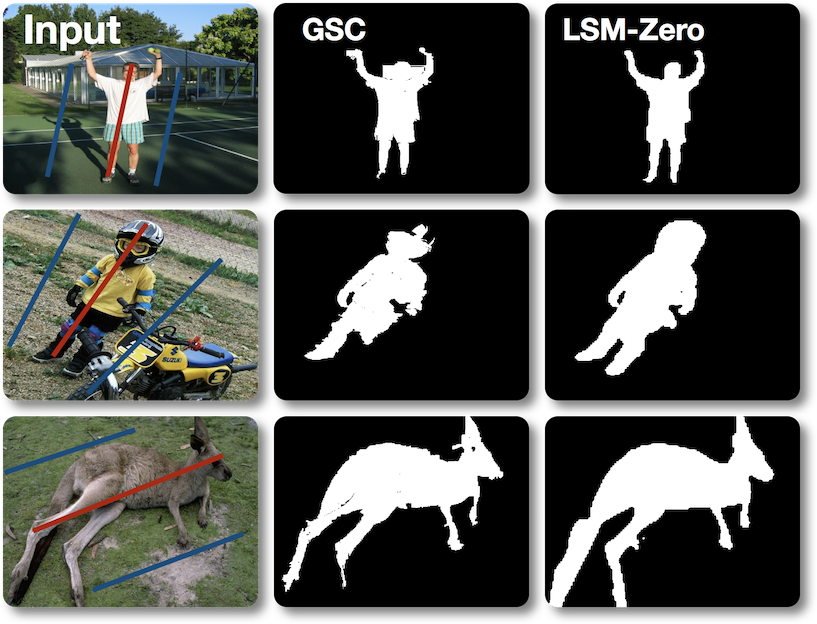}
	\vspace{-5pt}
	\caption{Our zero-shot generalized LSM model performs better than GSC~\citep{Gulshan10} for interactive segmentation.}
	\vspace{-5pt}
	\label{fig:int0}
\end{figure}

Similar to the other zero-shot generalization tests in Sec. 4.3, we also leave the interactive segmentation out for testing and train on the other tasks. When interact only once, the average IoU is 0.802 for our LSM model learned on the other tasks and tested on the interactive segmentation. Which is still superior than the conventional method~\citep{Gulshan10,RANDOMWALK} as shown in~\figref{int0}.


\end{document}